\documentclass[lettersize,journal]{IEEEtran}
\usepackage{amsmath,amsfonts}
\usepackage{algorithmic}
\usepackage{algorithm}
\usepackage{array}
\usepackage[caption=false,font=normalsize,labelfont=sf,textfont=sf]{subfig}
\usepackage{textcomp}
\usepackage{stfloats}
\usepackage{url}
\usepackage{verbatim}
\usepackage{graphicx}
\usepackage{cite}
\usepackage{bbding}
\usepackage[table]{xcolor}
\usepackage{siunitx}
\usepackage{multirow}
\usepackage{booktabs}
\usepackage{bm}
\begin{document}

\title{L-CLIPScore: a Lightweight Embedding-based Captioning Metric for Evaluating and Training}

\author{Li Li $^1$ \quad Yingzhe Peng$^{1}$ \quad Xu Yang$^1$*\thanks{*Corresponding author}   \quad Ruoxi Chen$^1$ \\ \quad Xu, Haiyang$^{2}$ \quad Yan, Ming$^{2}$ \quad 	
Huang, Fei$^{2}$\\
\normalsize$^1$ Key Laboratory of New Generation Artificial Intelligence Technology \& \\
\normalsize Its Interdisciplinary Applications, (Southeast University),Ministry of Education\\
\normalsize$^2$ Alibaba Group\\
\tt\small lilyli@seu.edu.cn, yinghzhe.peng@seu.edu.cn, xuyang\_palm@seu.edu.cn, 213200761@seu.edu.cn,\\
\tt\small shuofeng.xhy@alibaba-inc.com, ym119608@alibaba-inc.com
, f.huang@alibaba-inc.com}

\markboth{Journal of \LaTeX\ Class Files,~Vol.~14, No.~8, August~2021}%
{Shell \MakeLowercase{\textit{et al.}}: A Sample Article Using IEEEtran.cls for IEEE Journals}

\IEEEpubid{0000--0000/00\$00.00~\copyright~2021 IEEE}


\maketitle
\begin{abstract}
   We propose a novel embedding-based captioning metric termed as \textbf{L-CLIPScore} that can be used for efficiently evaluating caption quality and training captioning model. L-CLIPScore is calculated from a lightweight CLIP (\textbf{L-CLIP}), which is a dual-encoder architecture compressed and distilled from CLIP. To compress, we apply two powerful techniques which are weight multiplexing and matrix decomposition for reducing the parameters of encoders and word embedding matrix, respectively. To distill, we design a novel multi-modal Similarity Regulator (\textbf{SR}) loss to transfer more vision-language alignment knowledge. Specifically, SR loss amplifies the multi-modal embedding similarity if the given image-text pair is matched and diminishes the similarity if the pair is non-matched. By compressing and distilling by this novel SR loss, our L-CLIP achieves comparable multi-modal alignment ability to the original CLIP while it requires fewer computation resources and running time. We carry out exhaustive experiments to validate the efficiency and effectiveness of L-CLIPScore when using it as the judge to evaluate caption quality. We also discover that when using L-CLIPScore as the supervisor to train the captioning model, it should be mixed up by an n-gram-based metric and meanwhile analyze why using L-CLIPScore only will cause fail training. The code is given in \url{https://github.com/ForJadeForest/DistillCLIP}
\end{abstract}


\section{Introduction}
\begin{figure}
	\centering
	\includegraphics[width=0.5\textwidth]{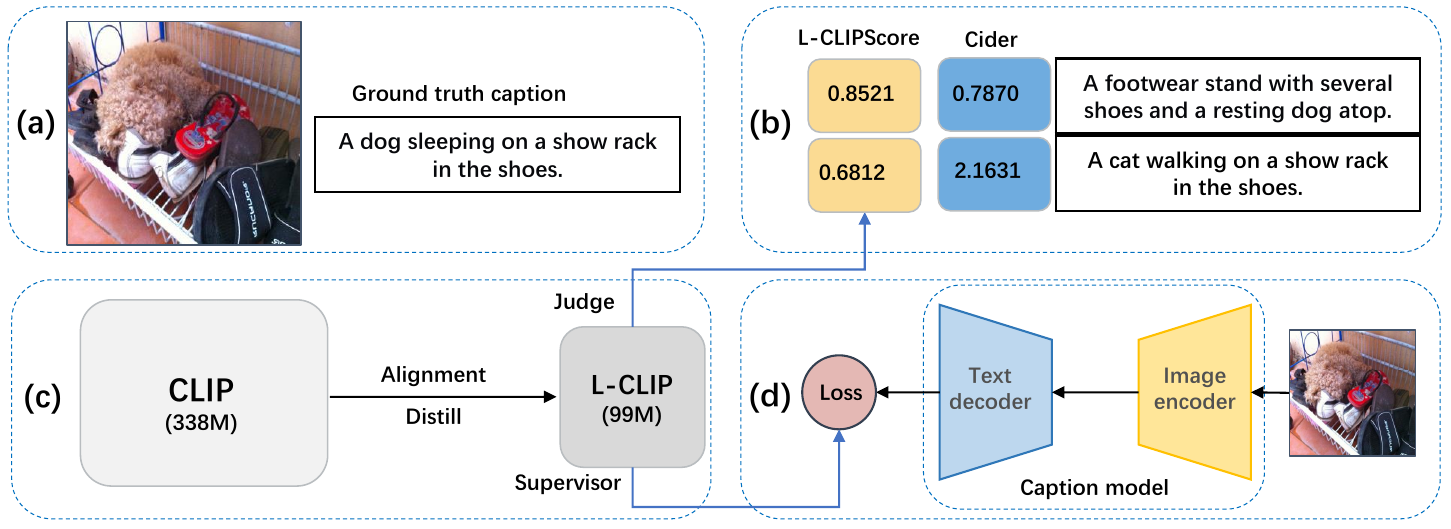}
	\caption[Figure]{(a) An image-caption pair. (b) Although the top caption is more matched with the image than the bottom one, CIDEr prefers the bottom one and our L-CLIPScore prefers the top one. (c) CLIP is compressed and the corresponding multi-modal alignment knowledge is also distilled into L-CLIP, which can be used as the judge to evaluate the caption in (b) and as the supervisor to train a captioning model in (d).}
	\label{fig:fig_intro}
\end{figure} 
Image Captioning, which aims at generating a sentence to exhaustively describe one source image, has achieved huge progress in the past few years from different perspectives, e.g.\ , the model architecture~\cite{soh2016learning, he2020image, cornia2020meshed,anderson2018bottom}, training objectives~\cite{rennie2017self}, or even datasets~\cite{sharma2018conceptual, lin2014microsoft},
Interestingly, only recently researchers discover that n-gram-based evaluation metrics, e.g.\ , BLEU~\cite{papineni2002bleu} or CIDEr~\cite{vedantam2015cider}, hinder the further development of image captioning~\cite{zhangbertscore} and some studies have highlighted challenges in text-image alignment evaluation\cite{yarom2023you,fastxx,lee2020vilbertscore,lifan}. Specifically, these n-gram-based metrics focus more on surface-form similarity instead of the semantic equivalence and thus fail to deal with the paraphrases, meaning-preserving lexical, and compositional diversity~\cite{zhangbertscore}. For example, Figure~\ref{fig:fig_intro}(b) shows that CIDEr prefers a worse caption (the top one) due to the surface-form similarity of ``in the shoes'' between the generated caption and the ground-truth caption.

To alleviate such limitation, researchers have begun exploring embedding-based metrics, e.g.\, BERTscore~\cite{zhangbertscore} uses BERT to embed the generated and human-annotated captions and then calculate the embedding similarities as the evaluation score. However, BERTscore still requires additional human-annotated captions as the n-gram-based metrics, while references are usually expensive and not available in some applications~\cite{lee2021umic}. To address such a problem, researchers~\cite{lei2021less,2019ViLBERT} exploit vision-language BERT-like Transformer models like CLIP and ViLBERT to embed the image and the generated caption for calculating the vision-language embedding similarities. Although the experiments demonstrate that these metrics achieve higher human correlations compared with n-gram-based ones, they still face one huge challenge which hinders more widely applications.

The challenge is that all these embedding-based models exploit the large-scale pre-training models, which are hardly used in cases where only lightweight devices are available or quick inferences are required. However, caption evaluation is frequently used in these cases. For example, caption evaluation helps visually impaired persons judge whether the generated caption is reliable~\cite{levinboim2021quality}, while carry-on medical devices usually only provide limited computation and storage resources. 
\IEEEpubidadjcol

Moreover, a metric can not only act as the judge to evaluate the quality of the generated captions but can also act as the supervisor to train captioning models. For example, CIDEr is widely used as the sentence-level reinforcement reward for training SOTA models~\cite{rennie2017self}. However, the previous studies about embedding-based metrics neglect this important function. Meanwhile, directly setting these large-scale BERT-based metrics as the loss module will largely increase the training computation burdens since they contain huge amounts of parameters. Thus a more lightweight embedding-based loss module is preferred. 

\IEEEpubidadjcol

Interestingly, researchers discover that large-scale BERTs are usually over-parameterized for solving some specific tasks~\cite{kovaleva2019revealing, zhang2022minivit}, which means that they can be compressed while keeping the performance for these tasks. Motivated by this, as shown in Figure.~\ref{fig:fig_intro}(c), we propose to compress CLIP and distill the multi-modal alignment knowledge of CLIP to get a \textbf{L}ightweight \textbf{CLIPScore} (\textbf{L-CLIPScore}) for efficient evaluation (Figure.~\ref{fig:fig_intro}(b)) and training (Figure.~\ref{fig:fig_intro}(d)). To achieve this, we apply two powerful compression strategies which are weight multiplexing~\cite{zhang2022minivit} and matrix decomposition~\cite{lan2019albert} for respectively reducing the parameters of encoder networks and word embedding matrix. Although some studies propose to distill CLIP for solving different tasks~\cite{dai2022enabling, wang2022clip}, they do not consider the multi-modal alignment but are more likely to use single-modal distillation strategies for separately distilling the vision and language encoders. Differently, we consider this by designing a multi-modal \textbf{Similarity Regulator (SR)} distillation loss, which will amplify the vision-language embedding similarity during distilling if the image-caption pair is matched and will diminish the similarity if the pair is non-matched. 
Our approach pioneeringly introduces contrastive learning into the knowledge distillation process from the teacher to the student model, enabling rapid transfer of the teacher's modality alignment knowledge. Compared to fine-tuning the student model with extensive image-text pairs to recover its multimodal capabilities, this method is more economical and efficient.

After compressing and distilling by our novel designed SR
loss, an efficient and effective L-CLIP is got for calculating L-CLIPScore. Specifically, the size of L-CLIP is only \textbf{99M} compared with the original \textbf{338M} of ViT-B/32 CLIP~\cite{radford2021learning} and the inference time decays from \textbf{225ms} to \textbf{124ms} running a 128-batch of image-caption paris. Meanwhile, as a judge for evaluating caption generation, its evaluation ability is still comparable with the original model in terms of the Kendall correlation, image-text matching accuracy, and the ability against the noise captions. Moreover, as a supervisor for training a generation model using the self-critical sequence training(SCST) strategy, we discover that only using L-CLIPScore will cause fail training and find that when mixing L-CLIPScore with n-gram-based metrics to train the model, the generated caption achieves higher quality in terms of both embedding-based and n-gram-based metrics. To sum up, our contributions are:

\begin{itemize}
	\setlength{\itemsep}{3pt}
	\setlength{\parsep}{3pt}
	\setlength{\parskip}{3pt}

	\item 
	we obtain a lightweight version of CLIP (L-CLIP) by applying model compression and distillation and develop L-CLIPScore, which can be used for efficient and effective evaluation and training.
	\vspace{-6pt}
	\item We propose a novel Similarity Regulator loss for distilling multi-modal alignment knowledge.
	\vspace{-6pt}
        \item We carry out exhaustive experiments to validate the effectiveness of L-CLIPScore acting as the judge and supervisor.
        \vspace{-6pt}
	\item We discover how to appropriately use L-CLIPScore as the loss module in self-critical sequence training.

\end{itemize}

\section{Related Work}

\subsection{Image Captioning}
Image Captioning has got a great success due to the development of various techniques. For example, the model architecture evolves from the classic CNN-LSTM pipeline~\cite{soh2016learning,2016SCA, 2017Knowing,he2020image,li2019TMMCNN,zhang2018high} to the modern popular pure Transformer pipeline~\cite{yu2019multimodal,cornia2020meshed,2022Compact,yan2021tasktf,shao2023textual} which is better at capturing the single- and dual-modal contexts. To better train these architectures, the training objective changes from the simple token-level cross-entropy loss to sentence-level reinforcement reward~\cite{rennie2017self}. The scale of captioning dataset also grows for feeding these stronger models, e.g.\ , from the elemental MSCOCO dataset~\cite{lin2014microsoft}, which contains 118K images and 590K captions annotated by experts, to modern ConceptNet, which collects 3.3M images and 3.3M captions from websites. Recently, researchers find that the widely used n-gram-based metrics, e.g.\ , BLEU, METEOR~\cite{banerjee2005meteor}, ROUGE~\cite{lin2004rouge}, SPICE~\cite{anderson2016spice}, CIDEr, hinder the further development of image captioning since they focus more on surface-form similarity instead of the semantic similarities~\cite{zhangbertscore}.

\subsection{Embedding-based Metrics}


Recent advances in text-image alignment evaluation have revealed significant challenges in handling complex semantic compositions through traditional embedding-based methods \cite{yarom2023you}. To enhance alignment capabilities, recent approaches focus on architectural innovations, such as lightweight Transformers for reduced complexity \cite{lifan} and memory-augmented embedding frameworks for fine-grained correspondence \cite{fastxx}. Concurrently, embedding-based metrics have emerged as powerful tools for image caption evaluation.
BERTScore~\cite{zhangbertscore} proposes embedding-based caption metrics by exploiting the large-scale pre-trained BERT~\cite{devlin2018bert}. In the beginning, only the similarities between the generated and human-annotated captions are considered. Then certain studies, which include CLIPScore~\cite{hessel2021clipscore}, ViLBERTScore~\cite{lee2020vilbertscore}, and Tiger~\cite{jiang2019tiger}, also consider directly measuring the similarity between the generated captions and the images to deal with the cases where no human-annotated captions are available. EMScore~\cite{shi2022emscore} further extends the captioning evaluation to the video case by considering both coarse- and fine-grained alignments. 

However, existing embedding-based metrics still rely on large-scale pretrained models, making them hardly usable for quick inference and efficient training. To amend this limitation, we propose to compress CLIP to get a lightweight CLIPScore, providing a more efficient embedding-based benchmark for both evaluation and training in image captioning task.


\subsection{Transformer Compression}
Transformer has gotten astonishing achievements in various fields including language~\cite{floridi2020gpt, lewis2020bart, devlin2018bert}, vision~\cite{liu2021swin, dosovitskiy2020image}, or even multi-modal domains~\cite{wang2022image, su2019vl, mu2022slip}. However, these models usually contain huge number of parameters, which makes them hardly be deployed into edge devices or achieve quick inference. Motivated by this, researcher propose various techniques like pruning~\cite{li2016pruning, zhu2021vision}, quantization~\cite{liu2021post}, and weight-sharing~\cite{lan2019albert, zhang2022minivit} to compress the vision or language Transformers. Here we use weight-multiplex and matrix decomposition to compress. Some studies also disill vision-language Transformers for solving diverse tasks, while most of them do not consider the multi-modal characteristic and only follow the single-modal methods~\cite{wu2022tinyvit, jiao2020tinybert, sanh2019distilbert, touvron2021training}, e.g.\ , distill attention matrices or the logits of the last layer to the student model. Compared with them, we consider the characteristic of caption evaluation and then propose a novel Similarity Regulator loss.

\section{L-CLIPScore}
In this section, we first introduce the model architecture of the Lightweight CLIP (\textbf{L-CLIP}) in Section~\ref{arch_lclip}. Then we present the multi-modal loss used during the distillation phase and show how to use the newly designed Similarity Regulator loss to transfer the multi-modal alignment ability from the teacher CLIP into the student L-CLIP (cf. Section~\ref{multi-modal-loss}). At last, we demonstrate how to calculate L-CLIPScore for using it as the judge to evaluate caption quality and as the supervisor to train captioning models (cf. Section~\ref{clip_as_supervisor}).

\subsection{Architecture of L-CLIP\label{arch_lclip}}
We first revisit the pipeline of CLIP~\cite{radford2021learning} and then introduce how to get our L-CLIP. Briefly, CLIP is a dual-encoder model that the language encoder is a Transformer~\cite{vaswani2017attention} and the vision encoder can be a vision Transformer following the original settings in~\cite{dosovitskiy2020vit} or a modified ResNet~\cite{he2019bag, zhang2019making}. These two single-modal encoders are trained by the vision-language contrastive loss on a website-collected dataset that contains 400 million image-text pairs. This contrastive loss pulls close the embeddings of the matched image-text pairs and pushes away the non-matched ones. In this way, CLIP learns the image-text alignment knowledge and thus be applied to measure whether a generated caption matches the given image~\cite{hessel2021clipscore}. 

\begin{figure}[t]
	\centering
	\includegraphics[width=0.4\textwidth]{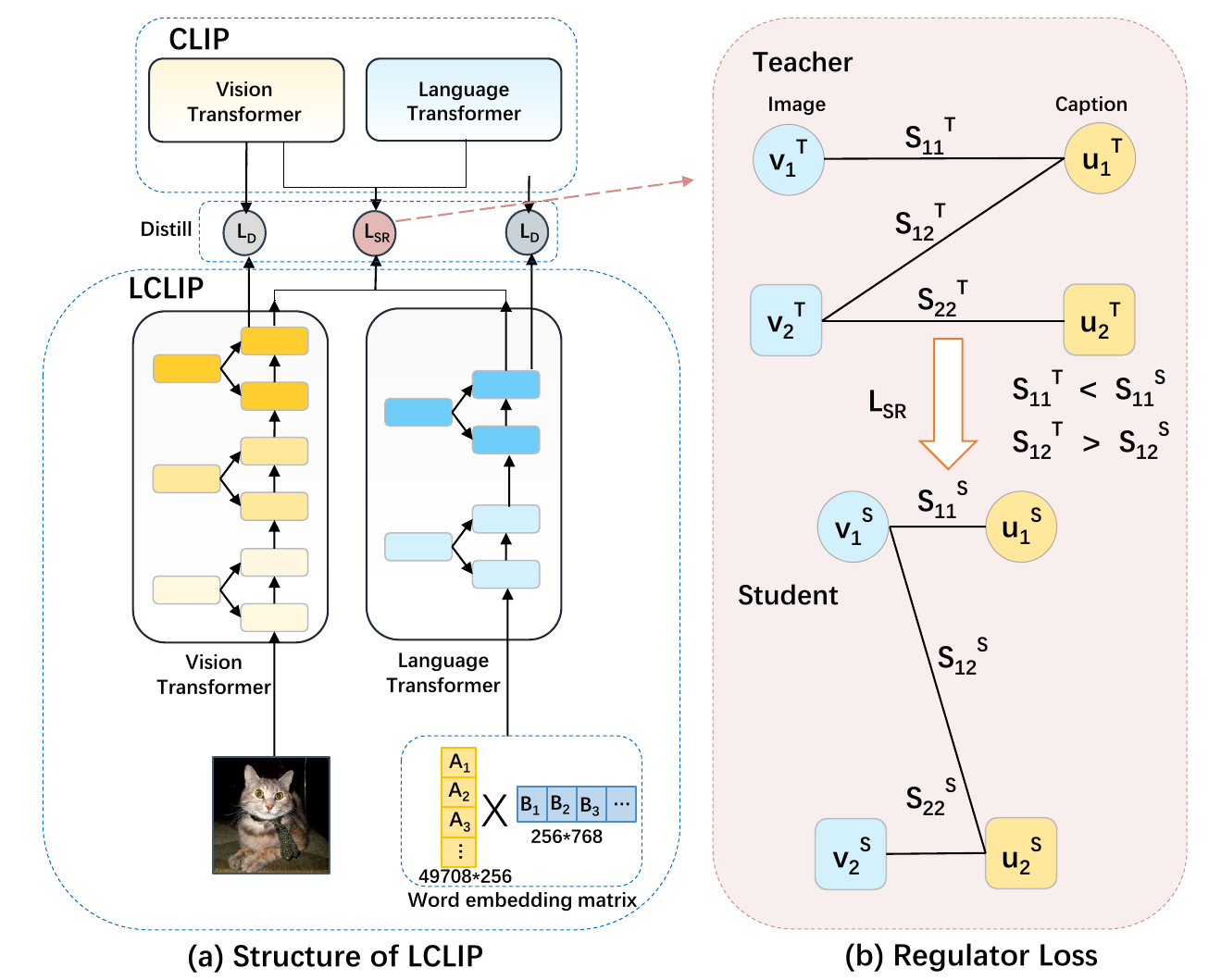}
	\caption[Figure]{(a) The architecture of L-CLIP. The vision/language encoders of L-CLIP contain 6/4 layers where each two Transformer blocks multiplex one shared block. The word embedding matrix decomposes into two small ones. The knowledge is distilled by single modal ($L_D$) and multi-modal Similarity Regulator Loss ($L_{SR}$). (b) The visualization of $L_{SR}$, which amplifies the multi-modal similarity if the image-caption pair is matched, e.g.\ ,  $s_{11}^T \leq s_{11}^S$, and diminishes the similarity if non-paired, e.g.\ , $s_{12}^T \geq s_{12}^S$. Note that we show the distance between two nodes, where a small distance denotes a larger similarity.} 
	\label{fig:fig_pipeline}
 \vspace{-0.4cm}
\end{figure}

However, the original CLIP, e.g.\ , the ViT-B/32 version, contains 338M parameters, and running it requires 1730-M GPU resources and 14-ms to deal with one image-text pair, which is hardly used in resource-limited devices for quick inference. Thus we propose to compress it to get a \textbf{L}ightweight \textbf{CLIP} for calculating image-text match score termed as \textbf{L-CLIPScore}. Specifically, as shown in Figure~\ref{fig:fig_pipeline}(a), L-CLIP is still a dual-encoder pipeline while both the vision and language encoders contain fewer parameters by applying weight multiplexing~\cite{zhang2022minivit} and matrix decomposition~\cite{lan2019albert}.

Specifically, both the vision and language Transformers are built by applying the weight multiplexing technique. This technique maps the weights of one learnable shared Transformer block, e.g.\ , the query/key/value transformation matrices or FFN weights, into a few diverse blocks by some learnable multiplex sub-networks. Then the weights of the shared Transformer block and the multiplex sub-networks are trained by Knowledge Distillation losses, which will be introduced later.  
As shown in Figure~\ref{fig:fig_pipeline}(a), the vision Transformer multiplexes 3 shared layers (the left yellow blocks in vision encoder) into 6 layers where each shared one is multiplexed into 2 diverse ones by 2 multiplexing sub-networks and the language Transformer multiplex 2 shared layers into 4 layers. 

Besides, for language Transformer, we use matrix decomposition~\cite{lan2019albert} to compress the word embedding matrix that accounts for a substantial proportion of the parameters. Specifically, as shown in bottom part of the language decoder in Figure~\ref{fig:fig_pipeline}(a), we decompose this $49408 \times 768$ embedding matrix, where $49408$ is the vocabulary size and $768$ is the dimension of word embedding, into two smaller $49408 \times 256$ and $256 \times 768$ matrices. This allows further compression of the text encoder without affecting the performance. 

After compressing, our L-CLIP contains 99M parameters (3.4 times smaller than CLIP) and it requires 980M GPU resources (1.76 times smaller) and 8ms (1.75 times quicker) to deal with one image-text pair. Given the architecture of L-CLIP, we next introduce how to distill knowledge from the ViT-B/32 CLIP.

\subsection{Multi-modal Distillation Loss\label{multi-modal-loss}}
To transfer the knowledge from the teacher CLIP to the student L-CLIP, we employ a two-stage distillation process.
Specifically, we first apply two single-modal distillation losses for respectively training the vision and language encoders. In this way, both vision and language encoders in L-CLIP learn to capture single-modal high-level semantic knowledge for preparing to learn the vision-language alignments. Let $\bm{V}=\{\bm{v}_0,...\bm{v}_N\}$ denote the vision embeddings and $\bm{U}=\{\bm{u}_0,...\bm{u}_M\}$ denote the text embeddings, where $\bm{v}_0$/$\bm{u}_0$ are respectively the vision/language global [CLS] tokens. For convenience, we use the superscript ``$T$/$S$'' to denote that these embeddings are output from the teacher/student model, e.g.\ , $\bm{V}^T$ denotes the vision embeddings from the teacher CLIP. 
Then two single-modal distillations have the same loss formula and we show the vision case here:
\begin{equation}
\small
\setlength{\abovedisplayskip}{2pt}
\setlength{\belowdisplayskip}{2pt}
    L_D = \frac{1}{2}|\bm{r}^T_0 - \bm{r}^S_0| + \frac{1}{2}(1 - \cos \left(\bm{r}^T_0, \bm{r}^S_0\right)), \\
    \label{equ:single-modal}
\end{equation}
where the $\bm{r}_0$ is either $\bm{v}_0$ or $\bm{u}_0$ and both the left and right terms enforce the teacher and student embeddings to be similar.

After single-modal distillation, we further consider multi-modal distillation. Although CLIP itself learns the image-text alignment knowledge, such knowledge is learned from the huge amount of samples and computation resources. Thus if we do not have the same scale data and computation resources during distilling, such alignment ability will be weakened. To amend this, we propose a novel Similarity Regulator loss to enhance the alignment ability during distilling. Specifically, as shown in Figure.~\ref{fig:fig_pipeline} (b), for the matched $i$-th image and $i$-th text in a batch, we hope to amplify the embedding similarity from teacher to student, i.e.\ , the student embedding similarity $\text{S}_{ii}^S$ should be larger than the teacher similarity $\text{S}_{ii}^T$. Meantime, for the non-matched $i$-th image and $j$-th text in a batch, we hope to diminish the embedding similarity, i.e.\ , $\text{S}_{ij}^S$ should be smaller than $\text{S}_{ij}^T$. In this way, we have the Similarity Regulator loss as:
\begin{equation}
\small
\setlength{\abovedisplayskip}{2pt}
\setlength{\belowdisplayskip}{2pt}
\begin{aligned}
    L_{SR}= & \sum_{i=1}^B\text{max}(0, \text{S}_{ii}^T-\text{S}_{ii}^S) + \\
    & \sum_{i=1}^B\sum_{j \ne i}^B\text{max}(0, \text{S}_{ij}^S-\text{S}_{ij}^T),
\end{aligned}
\label{equ:multi-modal}
\end{equation}
where $\text{S}_{ij}$ is set to the inner product of the global [CLS] tokens $\bm{v}_0$ of the $i$-th image and $\bm{u}_0$ of the $j$-th caption, and $B$ is the batch size. This loss naturally follows the characteristic of caption evaluation that the matched image-caption pairs should have similar vision language embeddings while the non-matched ones should have different embeddings.
Using the feature similarity of the teacher model as a reference for contrastive learning allows for the rapid extraction of knowledge acquired by the teacher model through extensive modality alignment training.
This design is particularly suitable for CLIP distillation scenarios, as it assesses how well captions match images. This type of distillation is specifically designed to achieve the goal of our paper, which is to construct an image captioning evaluation metric.
Additionally, we also employ Eq.~\eqref{equ:single-modal} to prevent the model from forgetting high-level semantic knowledge in the second stage.

\subsection{Usage of L-CLIPScore\label{clip_as_supervisor}}
After compressing by weight multiplex and matrix decomposition and distilling by Eq.~\eqref{equ:single-modal} and Eq.~\eqref{equ:multi-modal}, we get the L-CLIP that can calculate L-CLIPScore for two roles: as the judge to evaluate the caption quality and as the supervisor to train the captioning models.

\noindent\textbf{L-CLIPScore as the Judge.}
Given the original image and the generated caption, we respectively use the vision and language encoder of L-CLIP to get the vision embeddings $\bm{V}=\{\bm{v}_0,...\bm{v}_N\}$ and the word embeddings $\bm{U}=\{\bm{u}_0,...\bm{u}_N\}$. Then we follow~\cite{hessel2021clipscore} to calculate the L-CLIPScore as:
\begin{equation}
\small
\setlength{\abovedisplayskip}{2pt}
\setlength{\belowdisplayskip}{2pt}
    \text{L-CLIPS}(\bm{v_0}, \bm{u_0})=w * \max (\cos (\bm{v_0}, \bm{u_0}), 0)
\label{equ:l-clip-s}
\end{equation}
where $w$ is a re-scaling weight to stretch the similarity range from 0 to 1, we follow~\cite{hessel2021clipscore} to set $w=2.5$ since the originally calculated similarity ranges from 0 to 0.4. Also, if we have the human-annotated caption $\bm{U}^*=\{\bm{u}^*_0,...\bm{u}^*_N\}$, the reference version metric is:
\begin{equation}
\small
\setlength{\abovedisplayskip}{2pt}
\setlength{\belowdisplayskip}{2pt}
\begin{aligned}
& \text { RefL-CLIPS }(\bm{u_0}, \bm{U}^*, \bm{v_0})= \\
& \text { HM }(\text { L-CLIPS }(\bm{u_0}, \bm{v_0}), \max (\max _{\bm{r} \in \bm{U}^*} \cos (\bm{u_0}, \bm{r}), 0)
\end{aligned}
\label{equ:refl-clip-s}
\end{equation}
where HM is the harmonic mean.

\noindent\textbf{L-CLIPScore as the Supervisor.}
After the proposal of the self-critical training reward~\cite{luo2018discriminability}, n-gram metrics like CIDEr~\cite{vedantam2015cider} are widely used as the sentence-level training objectives. Since our L-CLIPScore can replace these n-gram metrics to judge the caption quality, it can also act as the supervisor for training. 

Given the vision embedding set $\bm{V}$, a captioning network $\mathcal{F}$ generates the caption $\bm{U}=\{\bm{u}_0,\bm{u}_1,...\bm{u}_M\}$ in an auto-regressive way:
\begin{equation}
\small
\setlength{\abovedisplayskip}{2pt}
\setlength{\belowdisplayskip}{2pt}
    P(\bm{u}_{t+1})=\mathcal{F}(\bm{V},\bm{U}_{0:t}),
\label{equ:at-regressive}
\end{equation}
where $P(\bm{u}_{t+1})$ is the predicted distribution of the $t+1$-th word and $\bm{U}_{0:t}=\{\bm{u}_0,...\bm{u}_t\}$ is the partially generated caption. To train this network, the token-level cross-entropy loss $L_{XE}$ and the sentence-level self-critical loss are in turn used as the supervision: 
\begin{equation}
\small
\setlength{\abovedisplayskip}{2pt}
\setlength{\belowdisplayskip}{2pt}
     L_{XE} =- \log P(\bm{U}^*),
 \label{equ:equ_celoss}
\end{equation}
\begin{equation}
\small
\setlength{\abovedisplayskip}{2pt}
\setlength{\belowdisplayskip}{2pt}
     L_{SC} = - \mathbb{E}_{\bm{U}^s \sim P(\bm{U})}(\mathbb{S}(\bm{U}^*, \bm{U}^s)),
 \label{equ:equ_rlloss}
\end{equation}
where $\bm{U}^*$ is the ground-truth caption, $\bm{U}^s$ is sampled from $P(\bm{U})$, $\mathbb{E}$ denotes the expectation, and $\mathbb{S}$ represents the sentence-level metric. To use L-CLIPScore as the supervisor, we can set $\mathbb{S}$ to Eq.~\eqref{equ:l-clip-s} when we only have source image and Eq.~\eqref{equ:refl-clip-s} when we also have reference captions.

\section{Experiments}
\subsection{Datasets}

\noindent\textbf{ImageNet-1M ~\cite{ILSVRC15}} with 1.2M images and \textbf{Conceptual caption~\cite{sharma2018conceptual}} with 3.3M captions are respectively used for single-modal distilling knowledge from teacher vision and language encoders to the student.

\noindent\textbf{MSCOCO~\cite{lin2014microsoft}} splits the whole dataset into $118,287/5,000/5,000$ train/val/test image-caption pairs. Each image in the training set contains five human-labelled captions. We use MSCOCO to test human correlations of L-CLIPScore and also use it to train caption models. 

\noindent\textbf{Flickr8K-Expert ~\cite{hodosh2013flickr}}, \textbf{Flickr8k-CF~\cite{hodosh2013flickr}} and \textbf{Composite dataset~\cite{aditya2015composite}} are used to calculate the kendall correlation coefficient $\tau$ of automatic metrics (n-gram-based or embedding-based metrics) comparing with the human rating. Flickr8K-Expert contains 5664 images and each image contains multiple captions. Each caption is scored by humans from 1 to 4 where 1 indicates that the image-caption pair is not matched and 4 denotes that they are highly matched. Flickr8k-CF has 48K image-text pairs where each pair is annotated by at least 3 humans. Each human provides a binary score where 1 indicates that the image and text are matched and 0 is denoted non-matched. The dataset averages these scores for each image-text pair as the final human rating. Composite contains 12K human judgments of images from MSCOCO, Flickr8k, and on Flickr30k. Each image with five references, one of the references is rated by humans on a scale of 1 to 5 from two dimensions(correctness and thoroughness).

\noindent\textbf{Pascal-50S\cite{vedantamconsensus}} includes 4K sentence pairs, all classified into four categories: HC, HI, HM, and MM. Each category uses different ways to generate two sentences. HC generates two human-annotated correct sentence pairs, HI contains one correct and one incorrect human-annotated sentence pair, HM contains one correct human-annotated sentence and one correct machine-generated sentence, and MM contains two machine-generated sentences. The human uses these 4K sentence pairs to compare with the 48 human-annotated references and selects one sentence from the pairs that are closer in meaning to the reference description as the labelled sentence.

\noindent\textbf{FOIL dataset\cite{shekhar2017foil}} consists of 32K images from MS-COCO2014 and it randomly modifies the noun phrases in the reference caption making the caption semantics wrong, i.e. FOIL caption. Each image has a (FOIL, true) pair of captions and expects metrics to distinguish true or FOIL. This helps us detect whether the L-CLIPScore will get the capability to discriminate inaccurate references.

\subsection{Implementation Details} \label{sec:4.2}
For CLIP, it contains a few CNN layers for converting image patches to dense representations, we directly copy them to the L-CLIP and freeze their weights during training. During the experiments, we have discovered that setting the dropout rate to a value other than 0 significantly affects the model's performance. Therefore, we set the dropout rate to 0. Additionally, we employ random enhancement for the image enhancement when using~Eq.\eqref{equ:multi-modal} to distill alignment knowledge. However, we do not use the enhancement strategy which may change the image semantics like changing the colors of images.

We choose ViT-B/32 as the teacher CLIP model. When using single-modal distillation losses, the learning rate is $5 \times 10^{-3}$ and the batch size is 4096. When using Similarity Regulator loss to transfer the multi-modal alignment knowledge, we freeze the language encoder to avoid it overfitting to the small dataset used in this stage. We set the learning rate to $1 \times 10^{-4}$ and the batch size to 2048. 
All hyperparameters are systematically validated on the validation set through controlled experiments, with final selections achieving optimal test performance.
All training processes are carried out with mixed precision and 4 RTX 3090 GPUs (24GB).

When using L-CLIPScore as the supervisor, we use the self-critical Transformer model\footnote{https://github.com/ruotianluo/self-critical.pytorch} as the captioning model. We train this model on the MS-COCO dataset using Karpathy training split and test on Karpathy test set\cite{2015Karpathy}. We utilize two pre-extracted visual features respectively, bottom-up features\cite{anderson2018bottom} and VinVL features\cite{2021VinVL}. This Transformer is firstly trained by cross-entropy loss Eq.~\eqref{equ:equ_celoss} 15 epochs and then by reinforcement reward Eq.~\eqref{equ:equ_rlloss} another 25 epochs. When using Eq.~\eqref{equ:equ_celoss}, we initialize the learning rate to \num{5e-4} and then decay it every three epochs by a factor of 0.8. When using Eq.~\eqref{equ:equ_rlloss}, we initialize the learning rate to \num{1e-5} and do not decay it. The ADAM\cite{2014Adam} optimizer is used during all the training stages. 

\subsection{L-CLIPScore as the Judge}
To test the effects of the applied techniques including compression strategies and the proposed multi-modal Similarity Regulator loss, we implement exhaustive ablation studies to test the effectiveness of L-CLIPScore acting as the judge to evaluate the caption quality.

\noindent\textbf{Ablation Studies.}
We deploy the following ablation studies. \textbf{CLIP}: The original CLIP is used. \textbf{WP\&Single}: We only use weight multiplexing to compress and only use single-modal distillation loss (Eq.~\eqref{equ:single-modal}) to distill. \textbf{WP\&SR}: Compared with WP\&Single, we also use the proposed Similarity Regulator loss (Eq.~\eqref{equ:multi-modal}). \textbf{WP+MD\&Single}: We use weight multiplexing and matrix decomposition to compress and only use Eq.~\eqref{equ:single-modal} to distill. \textbf{L-CLIP}: Compared with WP+MD\&Single, we also use SR loss to distill and then we get the proposed L-CLIP.

\noindent\textbf{Model Size and Inference Time.}
Table~\ref{model_size_speed} shows the model size and inference speed of various ablations. The running time is measured by running a batch of 128 in RTX3090. 
We can observe that our proposed Weight Multiplexing (WP) method reduces parameters by $1/2$ compared to the original model(CLIP) and improves inference speed by 100ms (nearly $1/2$). Additionally, combining WP with Matrix Decomposition (WP+MD) achieves a parameter reduction of $1/3$. It is also notable that matrix decomposition does not significantly impact model inference speed, likely because it introduces additional low-rank matrix multiplications.

\begin{table}[!t]
    \centering
    \caption{Model size and speed.}
        \begin{tabular}{lccc}
        \hline
            ~ & parameters & size(fp32) & mean time(ms)  \\ \hline
            CLIP & 151277312 & 338M & 225.96 \\ 
            WP & 76652416 & 147M  & 124.11  \\ 
            WP+MD & 51552896 & 99M & 124.9 \\ 
            \hline
        \end{tabular}

    \label{model_size_speed}
\end{table}

\noindent\textbf{Caption Evaluation.}
To measure the ability of caption evaluation, we follow~\cite{hessel2021clipscore} to calculate Kendall correlation, image-text matching accuracy, and the ability of against noise captions in Table~\ref{kendall-correlation}, ~\ref{pascal_ex}, and~\ref{foil}, respectively. Kendall correlation is a non-parametric statistic used to measure the strength of the association between two ranking vectors. In the experiments, given $M$ image-caption pairs, two $M$-dimensional ranking vectors are got. One of them is got from the human annotator by ranking the matching degree of these $M$ pairs and another one is got by an automatic metric like the n-gram-based or embedding-based metrics. This correlation ranges from -1 to 1 where a larger absolute value indicates a stronger association. The detailed computation formulas are given in the supplementary materials. For image-text matching accuracy, we calculate L-CLIPScore between one image and various captions and then select the caption with the highest score to see whether they are matched. We use the FOIL dataset to evaluate the ability of L-CLIPScore to identify incorrect or hallucinated captions by assessing their capacity to differentiate between FOIL and true captions. For reference-related metrics, we calculate the scores by using one reference (1ref-RefCLIPS) and four references (4ref-RefCLIPS).

\begin{table}[!t]
    \centering
    \caption{The Kendall correlation coefficient $\tau$ on Flickr8k-Expert, Flickr8k-CF, and Composite dataset. The grey background for CLIP denotes that this is a large teacher model.}
    \label{kendall-correlation}
    \scalebox{0.75}{
    \begin{tabular}{lccccc}
        \hline
        & Ref & Flickr8k \textbf{$\tau_c$} & Flickr30k \textbf{$\tau_b$} & Composite \textbf{$\tau_c$} \\
        \hline
        BLEU-4 & \Checkmark& 30.8 & 16.9 & 45.5 \\
        MEREOR & \Checkmark& 41.8 & 22.3 & 46.2 \\
        ROUGE & \Checkmark& 32.3 & 19.9 & 45.5 \\
        CIDEr & \Checkmark& 43.9 & 24.6 & 48.0 \\
        SPICE & \Checkmark& 44.9 & 24.4 & 49.9 \\
        \hline
        \rowcolor{gray!20}
        CLIP ~\cite{hessel2021clipscore} &\XSolidBrush& 51.2 & \textbf{34.4} & \textbf{53.8} \\
        WP\&Single &\XSolidBrush & 50.6 & 33.4 & 48.0\\
        WP\&SR &\XSolidBrush& 51.9 & 33.6 & 50.4 \\
        WP+MD\&Single &\XSolidBrush & 50.4 & 33.4 & 48.1 \\
        L-CLIP &\XSolidBrush & \textbf{51.6} & 33.5 & 50.3 \\
        \hline
        \rowcolor{gray!20}
        CLIP ~\cite{hessel2021clipscore}&\Checkmark & 53.0 & \textbf{36.4} & 55.4 \\
        WP\&Single&\Checkmark  & 52.9 & 36.0 & 56.4 \\
        WP\&SR &\Checkmark & 53.7 & 36.0 & \textbf{56.7} \\
        WP+MD\&Single&\Checkmark & 52.8 & 36.0 & 56.5  \\
        L-CLIP&\Checkmark & \textbf{53.8} & 36.0 & 56.5 \\
        \hline
    \end{tabular}
    }
\vspace{-0.2in}
\end{table}

\noindent\textbf{Comparing Methods.} We also compare the performance of n-gram-based metrics, e.g.\ , BLEU~\cite{papineni2002bleu}, METEOR~\cite{banerjee2005meteor}, ROUGE~\cite{lin2004rouge}, CIDEr~\cite{vedantam2015cider}, and SPICE~\cite{anderson2016spice} in Table~\ref{kendall-correlation} and~\ref{pascal_ex}. The previous proposed embedding-based metrics, e.g.\ , BERT-S~\cite{zhangbertscore}, TIGEr~\cite{jiang2019tiger}, ViLBERTScore~\cite{lee2020vilbertscore}, and BERT-S++~\cite{yi2020improving} are compared in Table~\ref{pascal_ex}. Note that the column ``Ref'' in these tables denotes that the reference captions are also used to calculate the scores.

\begin{table}[!t]
    \centering
    \caption{Pascal50S accuracy results with 5 references.}
    \label{pascal_ex}
\scalebox{0.7}{
    \begin{tabular}{llllll|l}
    \hline
        ~ &Ref & HC & HI & HM & MM & Mean \\ \hline
        length & ~ & 51.7 & 52.3 & 63.6 & 49.6 & 54.3 \\ 
        BLEU-4 ~\cite{papineni2002bleu} & \Checkmark & 60.4 & 90.6 & 84.9 & 54.7 & 72.6 \\ 
        SPICE ~\cite{anderson2016spice} & \Checkmark & 63.6 & 96.3 & 86.7 & 68.3 & 78.7 \\ 
        METEOR ~\cite{banerjee2005meteor} & \Checkmark & 63.8 & 97.7 & 93.7 & 65.4 & 80.1 \\ 
        ROUGE-L ~\cite{lin2004rouge}& \Checkmark  & 63.7 & 95.3 & 92.3 & 61.2 & 78.1 \\ 
        CIDEr ~\cite{vedantam2015cider}& \Checkmark  & 65.1 & 98.1 & 90.5 & 64.8 & 79.6 \\ \hline
        BERT-S (RoBERTa-F) ~\cite{zhangbertscore}& \Checkmark  & 65.4 & 96.2 & 93.3 & 61.4 & 79.1 \\ 
        TIGEr ~\cite{jiang2019tiger}& \Checkmark & 56.0 & 99.8 & 92.8 & 74.2 & 80.7 \\ 
        ViLBERTScore-F ~\cite{lee2020vilbertscore}& \Checkmark & 49.9 & 99.6 & 93.1 & 75.8 & 79.6 \\ 
        BERT-S++ ~\cite{yi2020improving}& \Checkmark  & 65.4 & 98.1 & 96.4 & 60.3 & 80.1 \\ 
        \hline
        \rowcolor{gray!20}
        CLIP ~\cite{hessel2021clipscore}& \XSolidBrush & 56.5 & \textbf{99.3} & \textbf{96.4} & 70.4 & 80.7 \\ 
        WP\&Single & \XSolidBrush & 54.6 & \textbf{99.3} & 94.3 & 73.5 & 80.4 \\ 
        WP\&SR& \XSolidBrush & 56.4 & 99.0 & 95.1 & 73.9 & 81.1 \\ 
        WP+MD\&Single & \XSolidBrush & 55.4 & 99.2 & 93.7 & 74.4 & 80.7 \\ 
        L-CLIP & \XSolidBrush & $\bm{57.0}$ & 99.1 & 94.7 & \textbf{74.6} & $\bm{81.4}$  \\
        \hline
        \rowcolor{gray!20}
        CLIP ~\cite{hessel2021clipscore}&\Checkmark & 64.5 & \textbf{99.6} & \textbf{95.4} & 72.8 & 83.1 \\ 
        WP\&Single &\Checkmark & 64.7 & 99.5 & 93.7 & 73.7 & 82.9 \\ 
        WP\&SR &\Checkmark& $\bm{66.1}$ & 99.5 & 95.0 & 74.4 & $\bm{83.8}$ \\ 
        WP+MD\&Single&\Checkmark & 64.2 & 99.5 & 93.9 & 74.1 & 83.0 \\ 
        L-CLIP &\Checkmark & 63.8 & \textbf{99.6} & 95.1 & \textbf{74.5} & 83.3 \\ 
        \hline
    \end{tabular}
    }
\end{table}

\begin{table}[!ht]
    \centering
    \caption{The accuracy in FOIL experiment. For the items marked with \scalebox{0.7}{\XSolidBrush}, we calculate the score using the equation~\eqref{equ:l-clip-s}, which does not use the references and thus the result is unchanged. }
    \label{foil}
    \begin{tabular}{lcccc}
    \hline
        ~ & ref & 1-ref & 4-ref & mean\\ \hline
        BLEU-4 &\Checkmark& 65.45 & 86.81 & 76.13 \\ 
        METEOR &\Checkmark & 69.89 & 82.13 & 76.01\\ 
        ROUGE &\Checkmark & 53.84 & 70.62 & 62.23\\ 
        CIDEr &\Checkmark & 85.67 & \textbf{94.25} & 89.96\\ 
        SPICE &\Checkmark & 61.35 & 80.62 & 70.99\\ \hline
        \rowcolor{gray!20}
        CLIP ~\cite{hessel2021clipscore} & \XSolidBrush & 88.72 & 88.72 & 88.72\\ 
        WP\&Single & \XSolidBrush & 86.12 & 86.12 & 86.12\\ 
        WP\&SR & \XSolidBrush & 89.45 & 89.45 & 89.45\\ 
        WP+MD\&Single & \XSolidBrush & 86.54 & 86.54 & 86.54\\ 
        L-CLIP & \XSolidBrush & $\bm{90.11}$ & $\bm{90.11}$ & $\bm{90.11}$\\ \hline
        \rowcolor{gray!20}
        CLIP ~\cite{hessel2021clipscore} &\Checkmark & 92.81 & 94.06 & 93.43\\
        WP\&Single &\Checkmark & 92.12 & 93.48 & 92.80\\ 
        WP\&SR &\Checkmark & 92.91 & 94.01 & 93.46\\ 
        WP+MD\&Single &\Checkmark & 92.06 & 93.43 & 92.75 \\ 
        L-CLIP &\Checkmark & $\bm{93.21}$ & $\bm{94.25}$ & $\bm{93.73}$\\ \hline
    \end{tabular}
\vspace{-0.2in}
\end{table}

\noindent\textbf{Results and Analyses.} 
Firstly, from Table~\ref{kendall-correlation}, ~\ref{pascal_ex}, and~\ref{foil}, by comparing the original teacher CLIP (CLIPScore) and the compressed CLIP using single-modal distillation loss (WP\&Single and WP+MD\&Single), we find that the compressed CLIP achieves lower performances in terms of all the scores since the compressed model has less parameters. However, the decay magnitude is not so large, e.g.\ , ``Mean'' in the 3-rd part of Table~\ref{pascal_ex} only decays 0.5 from CLIPScore to WP\&Single while the model size changes from 338M to 147M. When further applying matrix decomposition to decompose word embedding matrix(WP\&SR and L-CLIP) to further reduce the model size to 99M, certain performances is almost equal to the original CLIP, e.g.\ , by comparing ``Mean'' of WP+MD\&Single and CLIPScore in the 3-rd and 4-th parts of Table~\ref{pascal_ex}. Such observation corresponds to the results in~\cite{lan2019albert} that small embedding size can also get higher scores. Both comparisons validate the major motivation of this study: \textit{a lightweight CLIP can be got to efficiently and effectively evaluate the caption quality since the original CLIP is over-parameterized.} 

Next, we check the power of the newly proposed Similarity Regulator loss, we find that after using this loss, the evaluation ability increases, e.g.\ , in Table~\ref{kendall-correlation}, ~\ref{pascal_ex}, and~\ref{foil}, WP\&SR and L-CLIPScore are respectively better than WP\&Single and WP+MD\&Single. Moreover, L-CLIPScore can even outperform CLIPScore in certain metrics even if L-CLIPScore uses much fewer parameters. For example, in Table~\ref{pascal_ex}, ``Mean'' of L-CLIPScore in the 3-rd part achieves 0.7 higher scores than CLIPScore and in Table~\ref{foil} (81.4 vs. 80.7), ``Mean'' of L-CLIPScore in the 2-nd part achieves 1.39 higher scores than CLIPScore (90.11 vs. 88.72). These observations validate that: \textit{the proposed Similarity Regulator transfers more multi-modal alignment knowledge to the compressed L-CLIP for helping caption evaluation}. 

Besides, We also follow CLIP~\cite{hessel2021clipscore} to set the contrastive loss as the multi-modal distillation loss, but the model cannot converge well, e.g.\ , this contrastive loss only decreases a little and the image-text retrieval accuracy keeps decreasing during training. We guess the reason is that such contrastive loss may not work well if we do not have the same large-scale of training data as used to train CLIP. More details about this loss are given in the supplementary material.

Lastly, by comparing L-CLIPScore with n-gram-based metrics in Table~\ref{kendall-correlation}, ~\ref{pascal_ex}, and~\ref{foil}, we find that L-CLIPScore achieves higher scores in terms of human correlation, image-text matching ability, and the ability of identifying noise captions even when we do not use any human-labelled references. Once we use reference captions, L-CLIPScore achieves substantial improvements in these Tables compared with n-gram-based captions. Also, it can be found that L-CLIPScore achieves higher image-text matching ability than the other embedding-based metrics like ViLBERTScore even it uses a larger 1023M model\footnote{The model size comes from the link \url{https://github.com/facebookresearch/vilbert-multi-task}}. 

\begin{figure*}
	\centering
	\includegraphics[width=1.0\textwidth]{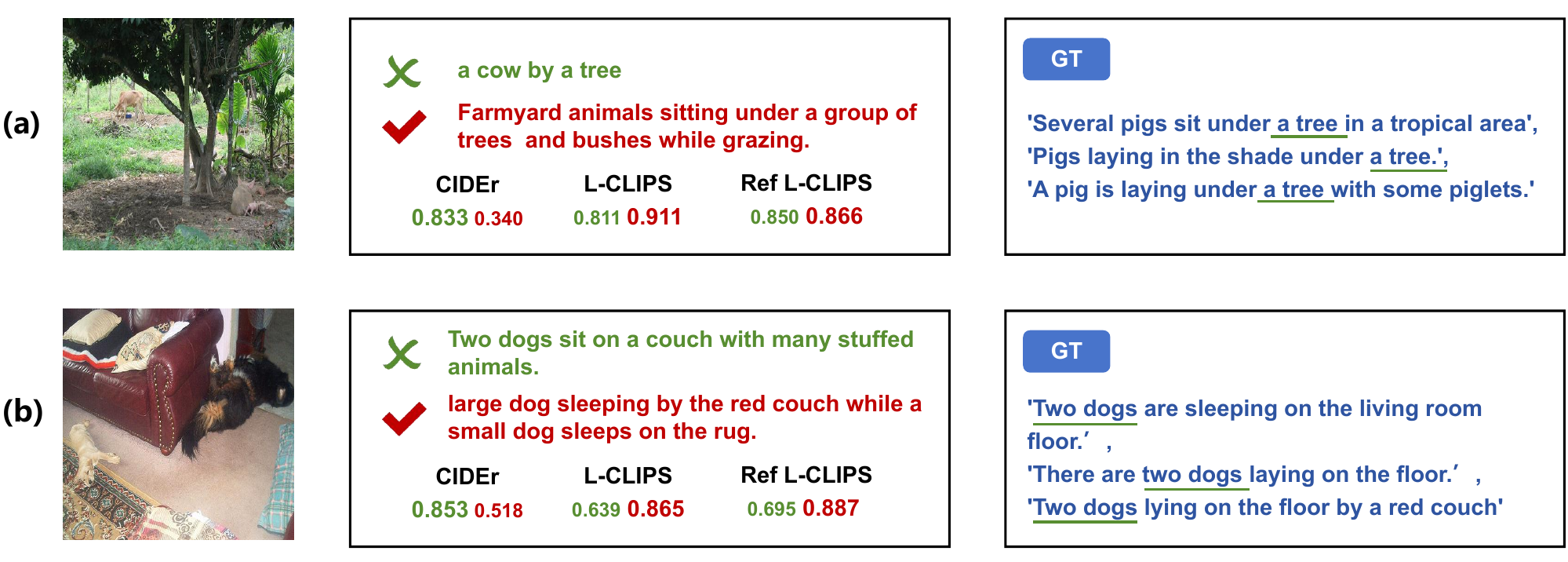}
	\caption[Figure]{Two visualized examples compare CIDEr and our L-CLIPScore. In each sub-figure, the green caption denotes the less matched one, the red one denotes the highly related one, and the blue captions denote the reference captions. The scores under each metric is also denoted by the corresponding colour, e.g.\ , green denotes the less matched one. We can find that due to certain surface similarities between the green captions with the blue ones, CIDEr returns high scores, e.g.\ , in (a), the word ``a tree'' spuriously improve the CIDEr score.} 
	\label{fig:fig_pascal}
\end{figure*}

We also visualize 2 qualitative examples from Pascal50S dataset to compare CIDEr and L-CLIPScore in Figure~\ref{fig:fig_pascal}. From these examples, we can find that CIDEr prefers the less matched caption while L-CLIPScore prefers the better one, e.g.\ , in (a), due to the high frequency of “a tree" in the reference, CIDEr returns a high score while L-CLIPScore does not, or in (b), the L-CLIPScore can assign a higher score to a more comprehensive and accurate sentence, making it more consistent with human preferences.

\subsection{L-CLIPScore as Supervisor}
\label{sec:supervisor}
To see whether L-CLIPScore can be used as the supervisor and what is a suitable strategy to use it, we set L-CLIPScore to the loss module to train the captioning model. It should be stressed that the motivation of these experiments is \textit{not to outperform the SOTA captioning models} but only to find the suitable way of using L-CLIPScore. Thus we do not use sophisticated model architecture and complex training strategy but only use the classic Transformer with a simple training strategy introduced in~\ref{sec:4.2}. 

\noindent\textbf{Comparing Methods.} We implement the following comparing methods. $\mathbb{S}_{\textbf{CD}}$: the sentence-level metric $\mathbb{S}$ in Eq.~\eqref{equ:equ_rlloss} is set to CIDEr. $\mathbb{S}_{\textbf{L-C}}$: $\mathbb{S}$ is set to Eq.~\eqref{equ:l-clip-s}.  $\mathbb{S}_{\textbf{RefL-C}}$: $\mathbb{S}$ is set to Eq.~\eqref{equ:refl-clip-s}.  $\mathbb{S}_{\textbf{Mix}}$: $\mathbb{S}$ is set to the weighted sum of CIDEr and Eq.~\eqref{equ:refl-clip-s} and two weights are both 0.5. The results are shown in Table~\ref{sc_results}. $\mathbb{S}_{\textbf{CD}}$(BU) indicates bottom-up features and (VL) indicates VinVL features.

\noindent\textbf{Ablation Studies.}
To rigorously validate the design of mixed reward, we have designed ablation studies to investigate the impact of different weights assigned to the weighted sum of the CIDEr and L-CLIPScore metrics.
Specifically,
\[
\mathbb{S}_{\textbf{Mix}} = \alpha \mathbb{S}_{\textbf{CD}} + (1 - \alpha) \mathbb{S}_{\textbf{L-C}},
\] 
we conducted systematic ablation experiments with 
\[
\alpha \in \{0, 0.3, 0.5, 0.7, 1.0\}.
\]
When $\alpha$ is 0, it is equivalent to $\mathbb{S}_{\textbf{CD}}$. When $\alpha$ is 1.0, it is equivalent to $\mathbb{S}_{\textbf{L-C}}$. When $\alpha$ is 0.5, it is equivalent to $\mathbb{S}_{\textbf{Mix}}$ in Table \ref{sc_results}.
The ablation results are shown in Table \ref{ablation}.

\begin{table}[!t]
    \centering
    \caption{The performance of using L-CLIPScore as the loss by using Bottom-Up (BU) and VinVL (VL) features.}
    \label{sc_results}
\resizebox{\linewidth}{!}{
    \begin{tabular}{lcccccc}
    \hline
        ~ & BLEU4 & METEOR & ROUGE-L & CIDEr & SPICE & L-CLIPScore\\ \hline
        $\mathbb{S}_{\textbf{CD}}$(BU) & 33.9 & 26.5 & 55.5 & 114.5 & 19.9 & 76.3\\
        $\mathbb{S}_{\textbf{L-C}}$(BU) & 18.3 & 24.1 & 45.8 & 47.2 & 18.2 & $\bm{82.3}$\\
        $\mathbb{S}_{\textbf{RefL-C}}$(BU) & 25.5 & 24.8 & 51.4 & 79.1 & 18.6 & 82.2\\
        $\mathbb{S}_{\textbf{Mix}}$(BU) & 35.4 & 27.0 & 56.5 & 118.7 & 20.6 & 77.7 \\
        \hline
         $\mathbb{S}_{\textbf{CD}}$(VL) & 33.9 & 26.7 & 56.0 & 117.1 & 20.3 & 76.5\\
        $\mathbb{S}_{\textbf{Mix}}$(VL) & $\bm{35.8}$ & $\bm{27.6}$ & $\bm{56.9}$ & $\bm{121.2}$ & $\bm{21.3}$ & 78.5\\ \hline
    \end{tabular}
    }
\vspace{-0.1in}
\end{table}

\noindent\textbf{Results and Analyses.}
\begin{figure*}
	\centering
	\includegraphics[width=1.0\textwidth]{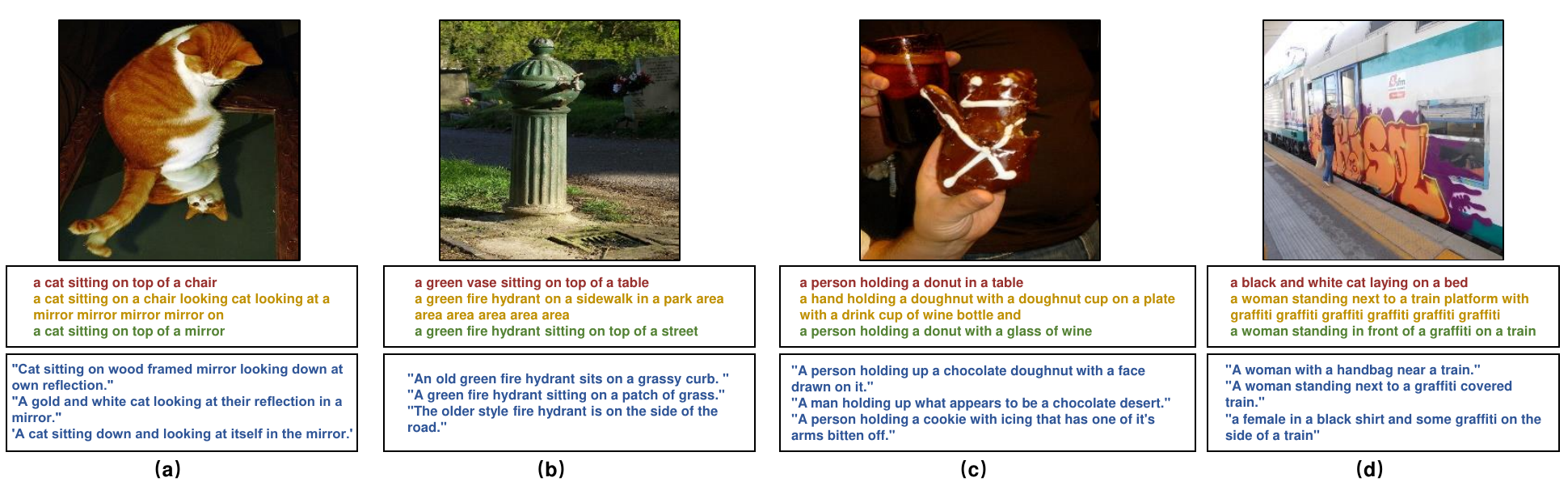}
	\caption[Figure]{We show 4 images with the captions. In the middle block, from top to bottom are the captions trained by CIDEr (red), L-CLIPScore (orange), and mixed reward (green). The last block shows the reference captions (blue).} 
	\label{fig:fig_supervisor}
\vspace{-0.1in}
\end{figure*} 

Interesting, from Table~\ref{sc_results}, we can find that the models solely trained by L-CLIPScore, e.g.\ , $\mathbb{S}_{\textbf{L-C}}$ and $\mathbb{S}_{\textbf{RefL-C}}$, can achieve higher L-CLIPScore while achieves much lower n-gram scores compared with $\mathbb{S}_{\textbf{CD}}$. 
Figure~\ref{fig:fig_supervisor} shows a few examples where the captions are generated by using different sentence-level rewards. We can find that when only using L-CLIPScore, the decoder prefers to produce numerous repeated visual tokens, causing the sentence structure to break down and the grammar to become chaotic. 
Meanwhile, utilizing pure CIDEr, $\mathbb{S}_{\textbf{CD}}$ achieves moderate linguistic quality(CIDEr) but exhibits relatively poor visual grounding(L-CLIPScore). 


This phenomenon aligns with theoretical insights about metric complementarity. 
Firstly, traditional n-gram metrics like CIDEr enforce syntactic fidelity by rewarding lexical overlap with references but struggle to assess visual-semantic alignment, often favoring safe, template-like phrases. 
Secondly, the observed language collapse under pure L-CLIPScore supervision stems from fundamental limitations of embedding-based metrics in reinforcement learning frameworks. 
In reward-driven training, models inherently tend toward reward hacking\cite{amodei2016rewardhacking} - optimizing for superficial metric patterns rather than genuine task objectives.
While L-CLIPScore effectively measures vision-language alignment through embedding similarity, it lacks explicit constraints on syntactic validity. This creates a pathological incentive structure: models discover they can artificially inflate L-CLIPScore by generating repetitive visual keywords that strongly correlate with image embeddings, while disregarding grammatical connectors (articles, prepositions) and sentence structure.
This phenomenon mirrors known challenges in RL-based text generation\cite{pan2024rewardhacking}.

Our mixed reward formulation $\mathbb{S}_{\textbf{Mix}}$ strategically counters this by retaining CIDEr’s n-gram constraints. 
L-CLIPScore guides vision-language alignment by measuring how captions capture latent visual patterns, while CIDEr the n-gram metric prevents the problem of repetitive words by forcing the model to follow the syntactic structure of the ground-truth references. 
As a result, the generated captions can be visually descriptive and linguistically fluent.
For example, Figure~\ref{fig:fig_supervisor} shows that the decoder utilizing $\mathbb{S}_{\textbf{Mix}}$ can quickly comprehend the concepts of “mirror" through supplementary image information, and also learn the uncommon noun entity “green fire hydrant." For image (c) in Figure~\ref{fig:fig_supervisor}, the $\mathbb{S}_{\textbf{Mix}}$ model describes the object “wine", which none of the reference captions mention.

\begin{table}[!t]
    \centering
    \caption{Ablation study on $\alpha$ values. Using bottom-up features\cite{anderson2018bottom}.}
    \label{ablation}
    \vspace{0.1in}
\resizebox{\linewidth}{!}{
    \begin{tabular}{lcccccc}
    \hline
        $\alpha$ & BLEU4 & METEOR & ROUGE-L & CIDEr & SPICE & L-CLIPScore\\ \hline
         0  & 33.9 & 26.5 & 55.5 & 114.5 & 19.9 & 76.3\\
         0.3 & 34.2 & 27.0 & 55.7 & 116.2 & 20.6 & 77.3\\
         0.5 & 34.4 & 26.7 & 55.8 & 115.3 & 20.4 & 77.4\\
         0.7 & $\bm{34.6}$ & $\bm{27.0}$ & $\bm{56.0}$ & $\bm{116.7}$ & $\bm{20.7}$ & 78.0\\
        1.0 & 18.3 & 24.1 & 45.8 & 47.2 & 18.2 & $\bm{82.3}$ \\
        \hline
    \end{tabular}
    }
\end{table}

Meanwhile, from the ablation results of several different ratios of mixed rewards in Table \ref{ablation}, we find that the performance differences among mixed rewards with different weights were not significant. This indicates that simply combining the two metrics as signals can yield good results. 
However, we also notice that while the model experiences language collapse when \(\alpha\)=1.0 (i.e., when trained solely with L-CLIPScore), the relatively optimal performance across nearly all metrics is achieved when the ratio of L-CLIPScore to CIDEr is 7:3. 
This may indicate that focusing more on visual accuracy while ensuring linguistic and grammatical constraints can lead to better model performance.
Additionally, simply incorporating a 0.3 weight of L-CLIPScore into the reward of SCST improves the model's performance by significant points. 
These two observations also demonstrate the effectiveness of introducing L-CLIPScore as a supervision signal. It helps the model better learn to leverage visual information for reasoning and highlights the advantages of using mixed rewards. Specifically, mixed rewards enhance visual accuracy while imposing constraints on language syntax. When sentence collapse occurs, the CIDEr score acts as a penalty, guiding the model to focus on grammatical correctness.

\section{Conclusion and Limitation}
In this paper, we propose to compress CLIP to get a Lightweight CLIP (L-CLIP) to calculate L-CLIPScore for efficiently and effectively evaluating caption quality and training caption models. To get the architecture of L-CLIP, we apply two compression techniques which are weight multiplexing and matrix decomposition to reduce the model size. We also distill the multi-modal alignment knowledge from the teacher CLIP to our student L-CLIP. To enhance the knowledge transfer, we propose a novel Similarity Regulator loss which amplifies the similarity of the matched image-text pair and diminishes the similarity of the non-matched pair. After compression, the model size changes from 338M to 99M and the running time decays from 225ms to 124ms when evaluating 128-batch image-caption pairs and meanwhile, the evaluation performance is still comparable to the original model. 

One limitation is shown in Sec~\ref{sec:supervisor} that if we directly use L-CLIPScore as the supervisor to train the captioning model, the model will make the model focus more on vision patterns while neglecting the linguistic constraint. As a result, the generated caption may repeat the same words about one vision object. By mixing L-CLIPScore and CIDEr, we solve this limitation in that the generated captions achieve better performance in terms of all the metrics. 

Another limitation of all the embedding-based metrics is that such metrics can not be used to measure the captioning model trained by the same embeddings. For example, if CLIP is used to evaluate the quality, then it will naturally return a high score to the captioning model that is also trained by using CLIP-based vision and language embeddings. To alleviate such model preference, in the future, we consider to distill a few more large-scale vision-language BERTs into one single lightweight model. In this way, the multi-modal alignment knowledge of the other VL-BERTs can be used to against the bias from the single VL-BERT.

\bibliographystyle{IEEEtran}
\bibliography{reference.bib}


 




\vfill

\end{document}